\documentclass[preprint,12pt,authoryear]{elsarticle}

\usepackage{amssymb}
\usepackage{amsmath}
\usepackage{graphicx}  
\usepackage{ulem}     

\newcommand{\maxval}[1]{\textbf{#1}}

\journal{Engineering Applications of Artificial Intelligence}

\begin{document}

\begin{frontmatter}

\title{Multi-Stage Patient Role-Playing Framework for Realistic Clinical Interactions} 

\author[1]{Shijie Jiang}
\author[1]{Zefan Zhang}
\author[2]{Kehua Zhu}
\author[1]{Tian Bai\corref{cor1}}
\author[3]{Ruihong Zhao\corref{cor2}}
\cortext[cor1]{Corresponding author: baitian@jlu.edu.cn (Tian Bai)}
\cortext[cor2]{Corresponding author: ruihongzhao8@jlu.edu.cn (Ruihong Zhao)}

\address[1]{College of Computer Science and Technology, Key Laboratory of Symbolic Computation and Knowledge Engineering, Ministry of Education, Jilin University, Changchun 130012, China}

\address[2]{College of Software, Jilin University, Changchun 130012, China}

\address[3]{Gastrointestinal Endoscopy Center, The First Hospital of Jilin University, Changchun 130012, China}






\begin{abstract}

The simulation of realistic clinical interactions plays a pivotal role in advancing clinical Large Language Models (LLMs) and supporting medical diagnostic education. Existing approaches and benchmarks rely on generic or LLM-generated dialogue data, which limits the authenticity and diversity of doctor-patient interactions. In this work, we propose the first Chinese patient simulation dataset (Ch-PatientSim), constructed from realistic clinical interaction scenarios to comprehensively evaluate the performance of models in emulating patient behavior. Patients are simulated based on a five-dimensional persona structure. To address issues of the persona class imbalance, a portion of the dataset is augmented using few-shot generation, followed by manual verification. We evaluate various state-of-the-art LLMs and find that most produce overly formal responses that lack individual personality. To address this limitation, we propose a training-free Multi-Stage Patient Role-Playing (MSPRP) framework, which decomposes interactions into three stages to ensure both personalization and realism in model responses. Experimental results demonstrate that our approach significantly improves model performance across multiple dimensions of patient simulation. Our dataset is available at https://github.com/SerajJon/MSPRP.

\end{abstract}



\begin{keyword}
Patient Role-Playing, Large Language Models, Clinical Simulation
\end{keyword}

\end{frontmatter}

\section{Introduction}

The rapid advancement of Large Language Models (LLMs) (\cite{grattafiori2024llama,hurst2024gpt,team2024gemini,yang2025qwen3}) has attracted widespread attention due to their remarkable capabilities in contextual understanding and semantic reasoning. Among their emerging applications, text-based Role-Playing Language Agents (RPLAs) stand out, enabling intelligent agents to emulate diverse personas and driving a wide range of applications, from digital clones and AI chatbots to role-playing games and social science research (\cite{wang2024rolellm,zhao2025medrag}), highlighting the growing integration of intelligent agents into daily life.

In the medical domain, role-playing in realistic clinical interactions is of particular importance. Expert systems that simulate doctors can provide users with extensive medical knowledge and diagnostic guidance, aiming to improve accuracy in tasks such as disease diagnosis, triage recommendation, and treatment plan generation. Simulating patient roles is equally critical. Patient simulations not only enable the evaluation of LLMs’ diagnostic reasoning when acting as doctors but also support medical students in learning and assessing diagnostic skills within authentic clinical scenarios.

\begin{figure*}[t]
  \centering
  \includegraphics[width=\textwidth]{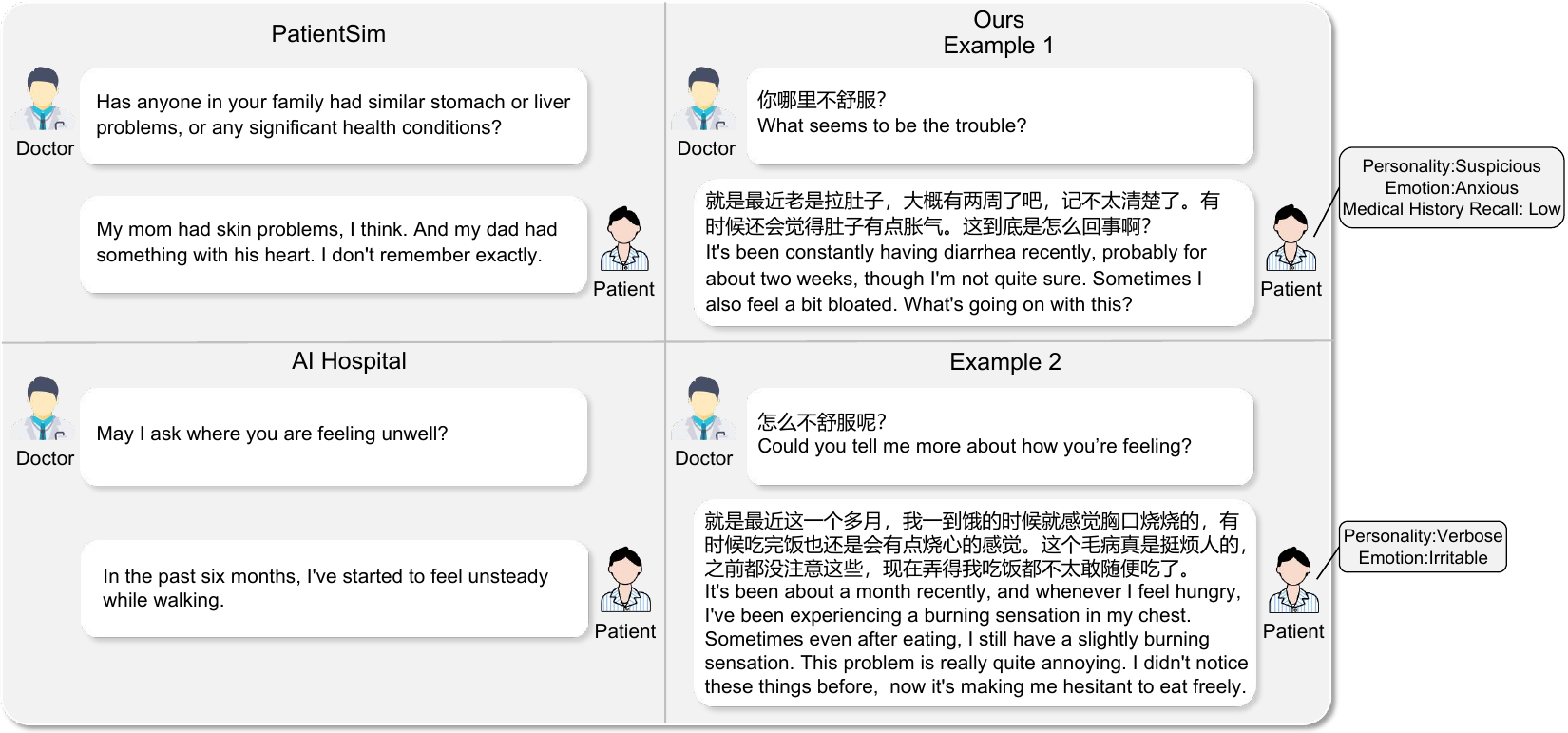} 
  \caption{Comparison of different patient simulation datasets (AI Hospital (\cite{fan2025ai}), PatientSim (\cite{kyung2025patientsim}) and ours).} 
  \label{ffirst+1} 
\end{figure*}

Recent advances in Large Language Models enable realistic simulation of clinical workflows. AI Hospital (\cite{fan2025ai}) evaluates LLM doctors on diagnostic accuracy, symptom collection, and test recommendations through multi-agent interactions with patients and examiners. EvoPatient (\cite{du2025llms}) and the AIE/SAPS framework (\cite{liao2024automatic}) generate high-quality, multi-turn doctor–patient dialogues using standardized or real hospital cases, while PATIENTSIM (\cite{kyung2025patientsim}) supports diverse patient types based on personality, language, and cognitive traits. These systems collectively demonstrate the potential of multi-agent frameworks for training and evaluating LLMs in realistic clinical scenarios. Despite recent efforts to create patient simulation datasets, most existing resources rely on LLM-generated data. These datasets are limited in their ability to capture authentic patient response styles, emotional expressions, and cognitive states, which restricts their applicability for advancing AI-assisted diagnostic tools and training, as the comparison shown in Figure \ref{ffirst+1}.

To tackle the challenges outlined above, this study presents the first patient simulation dataset built for authentic clinical dialogue scenarios. Focusing on gastroenterology, we collect outpatient records from 150 real patients, including medical histories, case reports, imaging findings, laboratory results, verbatim doctor–patient conversations, and others. To diversify patient profiles and balance category distribution, we further used LLMs with a few-shot strategy to generate additional cases, followed by thorough human verification. The final dataset contains 591 cases, each with roughly 10.5 dialogue turns on average. For evaluation, we prompt LLMs acting as simulated patients with real doctors’ questions, and use actual patient responses as ground truth. Model performance is assessed across five dimensions: basic content alignment, persona consistency, factual consistency, naturalness, and contextual relevance. These metrics ensure that simulated behaviors remain faithful to real clinical contexts. Across a range of LLMs, we observe a consistent pattern: most LLMs, regardless of scale, tend to produce overly uniform, formal, and emotionally flat responses during patient simulation. To mitigate this issue, we propose a multi-stage regulation framework that coordinates a base generation stage with control stages to stabilize and refine the simulated patient’s behavior. Experiments show that this framework yields substantial improvements across diverse LLM architectures.

Our contributions are summarized as follows:

\begin{itemize}
    \item We introduce Ch-PatientSim, the first benchmark built from real clinical doctor–patient interactions, designed to provide a reliable evaluation framework for LLM-based patient simulation.
    \item To address LLMs’ tendency to produce overly uniform, formal, and emotionally flat responses when simulating patients, we propose a multi-stage regulation framework that stabilizes and refines simulated patient responses.
    \item Our approach delivers significant improvements in both the realism and accuracy of patient simulation across multiple LLMs.
\end{itemize}

\section{Related work}

\subsection{Simulation of LLMs in Medical Scenarios}
With the advancement of Large Language Models (LLMs) in natural language understanding and generation, their applications in medical scenarios have gradually expanded from simple question-answering to complex agent simulation. Early research (\cite{zhou2025large,oniani2024enhancing,gaber2025evaluating,masanneck2024triage,zhao2025medrag}) on agents mainly focused on automating clinical processes, such as disease diagnosis, triage recommendation, or treatment plan generation, aiming primarily to improve task accuracy, with less attention paid to interaction realism.

Subsequently, some studies (\cite{li2024agent,fan2025ai,almansoori2025medagentsim}) attempted to simulate multi-role collaboration within hospitals through multi-agent systems, including doctors, nurses, and patients. These works demonstrated the potential of LLMs in information integration, reasoning, and collaborative decision-making, but patient agents are usually simplified as generic question-answering objects, lacking modeling of individual differences and language styles, which limit the realism of simulated interactions.

In recent years, some studies have shifted focus to doctor-patient dialogues and educational scenarios, evaluating the performance of doctor models in communication skills, empathy, and patient-centered interactions (\cite{chow2024ethical,zohny2025ai}), or exploring their potential as medical education aids (\cite{zhui2024ethical,lucas2024systematic}). These findings indicate that LLMs can perform multi-role tasks in medical scenarios, but making simulated patient roles semantically authentic and reflective of individual differences remains a central challenge.

\subsection{Role-Playing and Persona Modeling}
To enhance the naturalness and role consistency of LLMs in interactions, role-playing and persona modeling have become important directions. \cite{chen2024persona} pointed out in From Persona to Personalization that role-playing language agents, through explicit persona profiles or multi-attribute prompts, can improve consistency in identity maintenance, language style, and behavior logic. \cite{shao2023character} formalized role-playing as a trainable task, combining persona profiles with scene prompts to enhance role consistency and tone stability in language generation. \cite{peng2024quantifying} proposed a ``Global Faithfulness'' metric to quantify the alignment between generated behaviors and role definitions, and optimized strategies to improve the stability of persona-driven generation.

In terms of evaluation, \cite{tu2024charactereval} constructed a systematic Chinese role-playing benchmark, assessing model performance in conversational ability, character consistency, and role-playing attractiveness. \cite{zhou2025characterbench} further proposed a more comprehensive evaluation framework, covering believability, morality, memory, persona, knowledge, and emotion, providing standardized tests for persona-driven generation.

These studies suggest that LLMs can generate stable characters under personality control. However, in medical scenarios, due to strict semantic constraints and high ethical requirements, generation control methods that ensure medical accuracy while reflecting individualized traits are still needed.

\subsection{Personalized Patient Modeling in Medical Scenarios}

Driven by advances in general role-playing technologies, research on medical dialogue simulation is shifting from early process-level modeling toward more fine-grained persona modeling, integrating richer patient personality characteristics to improve interaction authenticity and diversity.

AI Hospital (\cite{fan2025ai}) constructs a dynamic multi-agent system of doctors, patients, and examiners, evaluating LLM doctors on symptom collection, test recommendation, and diagnosis using high-quality clinical records, though its primary focus remains on diagnostic workflows. EvoPatient (\cite{du2025llms}) introduces a collaborative multi-agent evolutionary framework that enables LLMs to autonomously simulate standardized patients and generate high-quality diagnostic dialogues through a dual doctor–patient agent setup. The AIE and SAPS simulators (\cite{liao2024automatic}) further enhance realism by grounding multi-turn interactions in 50 real clinical cases, with GPT-4 acting as both doctor and patient to produce approximately 10-round consultations. In addition, \cite{bodonhelyi2025modeling} examine how LLMs adapt communication strategies for different patient types (e.g., accusatory, dependent), highlighting the influence of emotion and communication style on interaction quality. \cite{kyung2025patientsim} extend this line of work with a persona-driven simulation framework based on multidimensional persona profiles, enabling more emotionally varied and personality-consistent dialogues and improving the realism of personalized patient interactions.

However, these studies mostly relied on qualitative personality type labels, lacking fine-grained, operationalized, structured dimensions to guide LLM generation precisely. This resulted in thin patient characterizations and inconsistent behaviors. Additionally, the doctor-patient dialogues in these studies are largely generated by LLMs, with limited real-world dialogue data to support them, restricting the medical authenticity of the patient language generated by the models.


\section{Dataset}

\begin{figure*}[t]
  \centering
  \includegraphics[width=\textwidth]{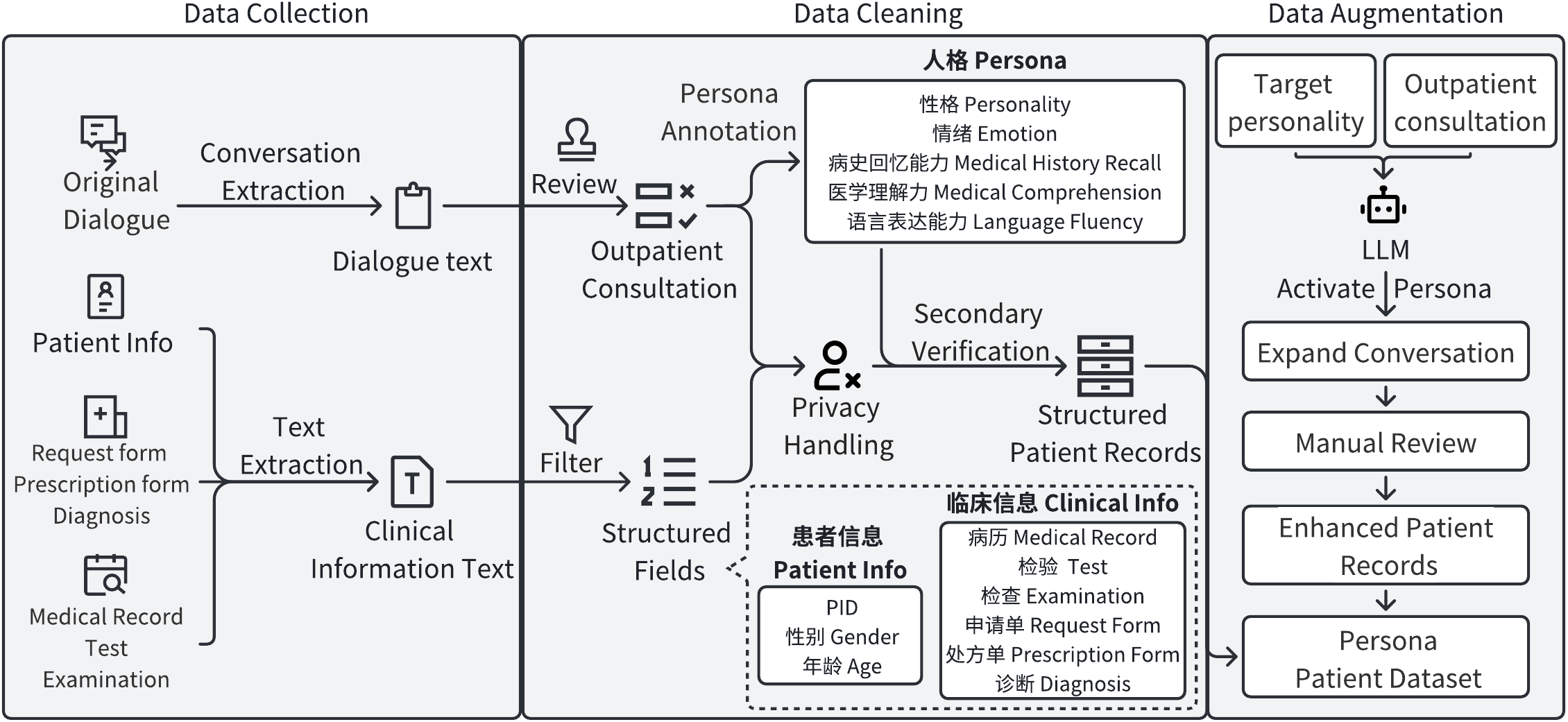} 
  \caption{The construction process of the Chinese patient simulation dataset (Ch-PatientSim).} 
  \label{ffirst_SHUJUJI} 
\end{figure*}

In this section, we focus on explaining the construction process of the Chinese patient simulation dataset (Ch-PatientSim), which is illustrated in Figure \ref{ffirst_SHUJUJI}.

\subsection{Data Collection}
The patient simulation dataset developed in this study is derived from real gastroenterology outpatient encounters in a hospital setting, encompassing multi-turn doctor–patient dialogues as well as complete medical record information. For each patient, the dataset includes basic demographic attributes, persona annotations, integrated patient clinical information, and outpatient conversation transcripts. Together, these components form a structured, multi-source corpus that supports the application of multi-dimensional control mechanisms during patient-simulation generation.

\subsection{Data Cleaning}
During data cleaning, medical professionals first review each consultation record sentence by sentence, distinguishing doctor and patient utterances and standardizing their formatting. Structured fields are then extracted from the electronic medical record system, with naming conventions unified and all identifiable information removed.

Persona attributes are annotated across multiple dimensions, including personality traits, emotional tendencies, recall ability regarding medical history, level of medical understanding, and expressive capability. After normalization and template-based cleaning, dialogue texts are integrated with the fused patient state information and persona labels to form standardized samples.

All data undergo two rounds of manual verification to ensure medical accuracy, linguistic naturalness, and consistency in persona expression. A third expert reviews cases showing annotation disagreement between the two annotators to determine the final outcome. Among the 591 samples, 43 cases require adjudication due to annotation inconsistencies.

\subsection{Data Augmentation}
Because the collected data exhibit category imbalance in persona and patient-state attributes, we apply LLM-based augmentation to rebalance the distribution. This process involves regenerating patient cases through persona reshaping of existing samples. A few-shot prompting strategy is used to guide LLM in producing additional dialogue records aligned with specific personas and state categories, thereby expanding the dataset while maintaining category balance. All augmented samples subsequently undergo both automated rule-based filtering and expert review to ensure medical correctness, persona consistency, and contextual coherence. Only dialogues that pass these checks are retained, guaranteeing that the expanded data remain reliable and clinically valid.

\subsection{Evaluation}

To quantitatively assess the semantic accuracy of model-generated patient responses, we employ a set of widely adopted automatic evaluation metrics, including BLEU, ROUGE, METEOR, and BERTScore. These metrics collectively examine different dimensions of semantic similarity between the generated response and the reference answer. Specifically, BLEU and ROUGE capture n-gram level correspondence, measuring lexical overlap and phrase-level fidelity; METEOR incorporates synonym matching and stemming to evaluate semantic alignment more flexibly; and BERTScore leverages contextualized embeddings to estimate deeper semantic similarity that goes beyond surface-level token matching. Together, these metrics provide a comprehensive examination of the model’s ability to produce semantically aligned and content-faithful outputs.

In addition, we adopt a model-based assessment framework using a strong LLM evaluator to judge higher-level pragmatic and contextual qualities of the generated responses. Following a discrete scoring scale from 1 to 5 (poor to excellent), we design carefully curated evaluation prompts and the collected patient responses to elicit consistent and interpretable judgments across four key dimensions:

Persona Consistency, which assesses whether the generated response adheres to the target patient persona, including personality, emotion, medical history recall, medical comprehension, and language fluency;

Factual Consistency, which measures the correctness of medical information and the presence of hallucinations or contradictory statements;

Naturalness, which evaluates the expressive fluidity, coherence, and human-likeness of the response in clinical communication;

Contextual Relevance, which examines whether the model appropriately responds to the doctor’s questions, maintaining situational awareness and progression within the conversation.

The combination of all these evaluation metrics enables a multi-angle and fine-grained assessment of both semantic fidelity and pragmatic communication quality, ensuring a robust understanding of the model’s performance in medical dialogue settings.

\subsection{Dataset Analysis}

\begin{figure*}[t]
  \centering
  \includegraphics[width=\textwidth]{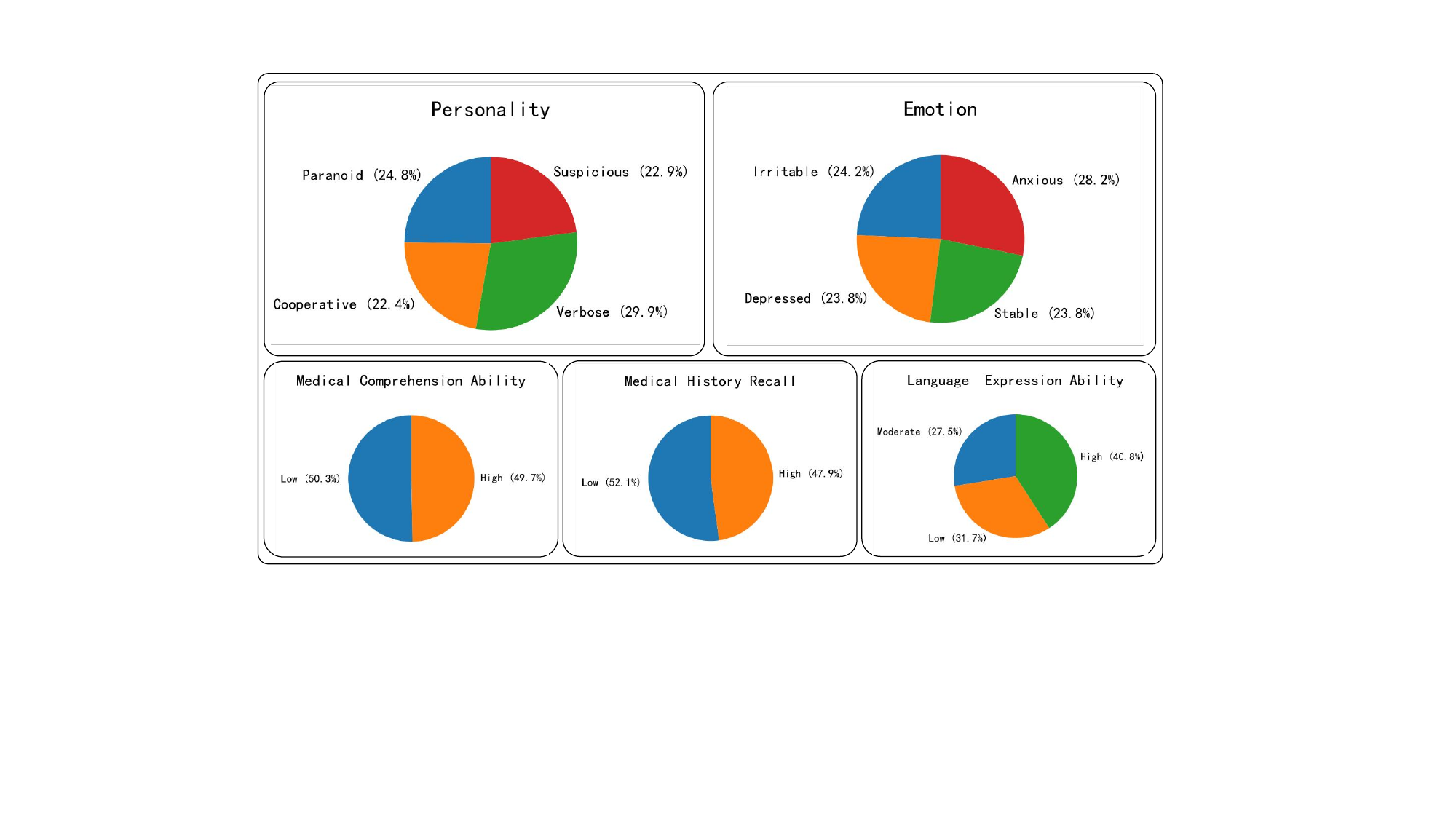} 
  \caption{Analysis of the persona attributes in the Ch-PatientSim dataset.} 
  \label{ffirst-bing} 
\end{figure*}

Statistical analysis shows that the dataset comprises 591 patients and 5,935 dialogue samples, covering a broad range of common clinical scenarios and diverse combinations of five-dimensional persona attributes. Each patient contributes an average of 10.5 dialogue turns, and the distribution across all five persona dimensions is well balanced, as illustrated in Figure \ref{ffirst-bing}, ensuring both diversity and representativeness of the training data. Grounded in authentic clinical content and medically sound annotations, the dataset provides robust support for patient-simulation research based on multi-dimensional coordinated control mechanisms. It offers a reliable data foundation for advancing the application of large language models in medical consultation dialogue generation.

\section{Method}

\subsection{Overall Framework}
\begin{figure*}[t]
  \centering
  \includegraphics[width=\textwidth]{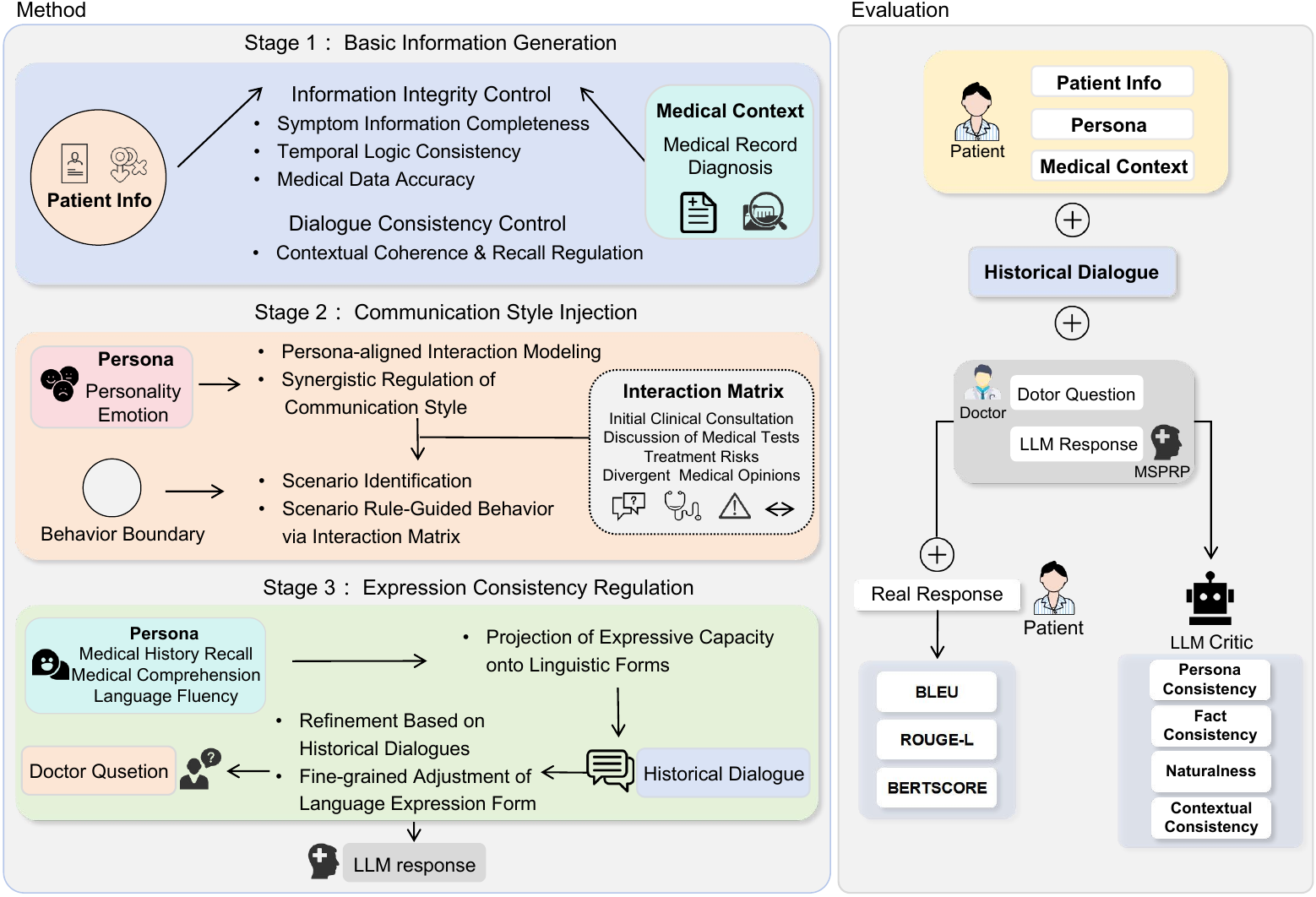} 
  \caption{Multi-Stage Patient Role-Playing (MSPRP) framework.} 
  \label{ffFirst} 
\end{figure*}

To address LLMs’ tendency to produce overly uniform, formal, and emotionally flat responses when simulating patients, we propose a multi-stage regulation framework that stabilizes and refines simulated patient responses. The method targets patient simulation within clinical interactions and centers on a persona modeling architecture consisting of five complementary dimensions organized into two functional categories, combined with a three-stage collaborative regulation mechanism. The overall framework is illustrated in Figure \ref{ffFirst}.


The objective of this task is to generate an answer $A$ based on several types of input information. These inputs include the patient's persona vector profile $P$, the Medical Context $C$ (i.e., the patient information, medical record, and diagnosis), the dialogue history $H$, and the current doctor's question $Q$. 

\subsection{Five-dimensional Persona}

The persona vector profile $P$ is defined as $
P = 
\bigl[
p^{\text{Personality}},\ 
p^{\text{Emotion}},\ \\
p^{\text{Medical History Recall}},\ 
p^{\text{Medical Comprehension}},\ 
p^{\text{Language Fluency}}
\bigr]
$, which could be divided into two functional categories: 
\begin{itemize}
    \item \textbf{Communication Style.} 
 The dimensions of personality and emotion characterize the patient’s stable interpersonal tendencies and affective tone during medical encounters.  
    They determine the patient’s communication attitude, response patterns, and overall interaction manner.

    \item \textbf{Expressive Capacity.} 
    The medical history recall, medical comprehension, and language fluency dimensions describe the patient’s ability to produce medically relevant information, understand clinical concepts, and articulate symptoms with appropriate detail.
\end{itemize}

These two categories play a crucial role by providing structured behavioral signals to the three-stage regulation mechanism, which allows the model to generate dialogue responses that are consistent, tailored to the individual, and medically appropriate.

\subsection{Three-stage Collaborative Regulation Mechanism}

To integrate patient personas into the dialogue generation process, we develop a three-stage collaborative regulation mechanism. The framework progressively ensures medical accuracy, persona alignment, and expression-level consistency.

\subsubsection{Stage 1: Basic Information Generation}

This stage focuses on establishing a medically reliable and logically coherent foundation for generation. It regulates the integrity and consistency of clinical information through:

\begin{itemize}
    \item Symptom information completeness: ensuring that the core symptom elements are fully preserved;
    \item Temporal logic consistency: preventing contradictions across the clinical timeline;
    \item Medical detail accuracy: covering medications, examinations, and diagnostic information;
    \item Contextual coherence and recall regulation: maintaining multi-turn factual consistency and modulating the extent of remembered information according to the medical history recall. 
\end{itemize}

Together, these controls ensure that subsequent persona-related adjustments are grounded on structurally sound medical content.

\subsubsection{Stage 2: Communication Style Injection}

After factual correctness is secured, this stage introduces the patient's communication style by aligning the generation with persona definitions and interaction patterns. It includes:

\begin{itemize}
    \item Persona-aligned interaction modeling: ensuring that the generated responses reflect the patient’s multi-dimensional persona;
    \item Synergistic regulation of communication style: capturing the joint influence of patients' personality and emotion on conversational behavior;
    \item Scenario identification: determining the clinical situation of the current turn;
    \item Scenario rule-guided behavior via an Interaction Matrix: categorizing common medical scenarios and prescribing personality–emotion behavioral rules for each situation.
\end{itemize}

Through these mechanisms, Stage 2 shapes how the patient speaks and reacts in a clinically coherent and persona-consistent manner.

\subsubsection{Stage 3: Expression Consistency Regulation}

This stage refines how persona traits are realized in linguistic form, mainly guided by the last three persona dimensions: medical history recall ability, medical comprehension ability, and language fluency.

\begin{itemize}
    \item Projection of expressive capacity onto linguistic forms: ensuring that expressive behavior aligns with the patient’s cognitive and communicative capacities;
    \item Refinement based on historical dialogues: adjusting expressions to maintain multi-turn stylistic coherence;
    \item Fine-grained adjustment of expressive form: controlling detail level, fluency, and clarity based on the patient’s expressive capacity.
\end{itemize}

Stage 3 ensures that the patient not only behaves according to their persona but also communicates in a way that matches their cognitive and linguistic profile.





\begin{table}
\centering
\caption{Basic evaluation of different LLMs on Ch-PatientSim.}
\vspace{3pt}
\resizebox{\textwidth}{!}{
\begin{tabular}{lccccccc}
\hline
Model & BLEU-1 & BLEU-2 & BLEU-3 & BLEU-4 & ROUGE-L & METEOR & BERTScore \\
\hline
Qwen2.5-7B          & 0.1828 & 0.0856 & 0.0504 & 0.0329 & 0.2051 & 0.2004 & 0.6437 \\
    GLM-4-9B            & 0.1824 & 0.0874 & 0.0521 & 0.0346 & 0.2099 & 0.1949 & 0.6478 \\
    Llama-3.1-8B        & 0.1656 & 0.0724 & 0.0422 & 0.0278 & 0.1839 & 0.1663 & 0.6293 \\
    Internlm3-8B        & 0.1552 & 0.0749 & 0.0454 & 0.0308 & 0.1844 & 0.1812 & 0.6299 \\
    Qwen2.5-72B         & 0.2006 & \uline{0.1019} & \uline{0.0634} & \uline{0.0431} & \uline{0.2257} & \uline{0.2256} & \uline{0.6551} \\
    GPT-4omini           & 0.1606 & 0.0738 & 0.0434 & 0.0285 & 0.1873 & 0.1940 & 0.6383 \\
    DeepSeek-V3         & 0.1348 & 0.0596 & 0.0335 & 0.0214 & 0.1606 & 0.1948 & 0.6266 \\
    Qwen2.5-7B+MSPRP    & \uline{0.2031} & 0.0956 & 0.0569 & 0.0375 & 0.2196 & 0.2041 & 0.6491 \\
    Qwen2.5-72B+MSPRP   & \maxval{0.2054} & \maxval{0.1057} & \maxval{0.0659} & \maxval{0.0450} & \maxval{0.2291} & \maxval{0.2313} & \maxval{0.6575} \\
\hline
\end{tabular}
}
\label{tab:llm_metrics1}
\end{table}

Automatic evaluation metrics for different LLMs. Missing entries indicate unavailable results.

\section{Experiments}

In this section, we conduct comprehensive experiments on the Ch-PatientSim dataset. Our overarching goal is to answer the following research questions:
\begin{itemize}
    \item RQ1: How well do current LLMs perform in patient simulation tasks?
    \item RQ2: How effective is the MSPRP method proposed in this paper?
    \item RQ3: Are LLMs’ patient-simulation behaviors natural and clinically plausible in realistic doctor–patient communication scenarios?
\end{itemize}

\subsection{Implementation Details}

We evaluate the role-playing capabilities of several LLMs, including Qwen2.5 (\cite{yang2025qwen3}), GLM-4 (\cite{glm2024chatglm}), DeepSeek-V3 (\cite{liu2024deepseek}), Llama-3.1 (\cite{grattafiori2024llama}), Internlm3 (\cite{cai2024internlm2}), and GPT-4o-mini (\cite{hurst2024gpt}). The persona evaluation metrics are tested using Qwen2.5-14B. The data augmentation method employs Qwen2.5-72B. All experiments are conducted on NVIDIA A40 GPUs.

\begin{table}
\centering
\caption{Persona evaluation of different LLMs on Ch-PatientSim.}
\vspace{3pt}
\resizebox{\textwidth}{!}{%
\begin{tabular}{lcccc}
\hline
Model & Persona Consistency & Factual Consistency & Naturalness & Contextual Relevance \\
\hline
 Qwen2.5-7B          & 3.748 & 3.795 & 3.824 & 3.780 \\
    GLM-4-9B            & 3.770 & 3.790 & 3.857 & 3.796 \\
    Llama-3.1-8B        & 3.416 & 3.515 & 3.523 & 3.384 \\
    Internlm3-8B        & 3.589 & 3.642 & 3.602 & 3.572 \\
    Qwen2.5-72B         & 3.870 & 3.896 & 3.914 & 3.910 \\
    GPT-4omini           & 3.858 & 3.855 & 3.906 & 3.871 \\
    DeepSeek-V3         & \uline{3.936} & 3.883 & \uline{3.962} & 3.916 \\
    Qwen2.5-7B+MSPRP    & 3.905 & \uline{3.918} & 3.942 & \uline{3.937} \\
    Qwen2.5-72B+MSPRP   & \maxval{3.939 } & \maxval{3.956} & \maxval{3.970} & \maxval{3.969} \\
\hline
\end{tabular}%
}
\label{tab:llm_human_eval1}
\end{table}

\subsection{RQ1: How well do current LLMs perform in patient simulation tasks?}

To answer RQ1, we evaluate a range of representative LLMs on the Ch-PatientSim dataset using both automatic similarity metrics and model-based human-aligned evaluations. The results in Table \ref{tab:llm_metrics1} and Table \ref{tab:llm_human_eval1} reveal clear and consistent trends across models.

First, as shown in Table \ref{tab:llm_metrics1}, Although the Qwen 2.5-72B performed best,across all models, BLEU, ROUGE, METEOR, and BERTScore remain relatively low, indicating that patient responses generated by current LLMs still diverge considerably from real patient utterances at the lexical and semantic levels. This highlights the inherent difficulty of matching the fragmented, colloquial, and personally varied speech patterns commonly observed in real clinical encounters.

Second, as presented in Table \ref{tab:llm_human_eval1}, top-performing models — including Qwen2.5-72B, GPT-4omini, and DeepSeek-V3 — achieve relatively high scores in persona and pragmatic evaluation, with DeepSeek-V3 distinguishing itself particularly in Persona Consistency and Naturalness. In contrast, smaller-scale models such as Llama-3.1-8B and Internlm3-8B exhibit notably lower performance across these dimensions. Importantly, no model surpasses a score of 4.0 in any evaluation category—a finding that underscores a critical limitation:  even large-scale models  struggle to consistently embody individualized patient traits, often failing to maintain a stable personality, coherent emotional tone, and consistent expression style in dialogues.

Third, a common issue across models is over-formalization: generated patient responses tend to be polite, structured, and medically coherent, but lack the diverse personalities, emotional fluctuation, and idiosyncratic expression patterns that characterize real patient interactions. This aligns with earlier observations that LLMs default to normative and institutionally ``safe'' tones when simulating patients.

Overall, these findings demonstrate that existing LLMs exhibit reasonable medical correctness and contextual logic, but fall short in simulating naturally varied and persona-driven patient behavior, which is the main gap that our proposed MSPRP framework aims to address.

\subsection{RQ2: How effective is the MSPRP method proposed in this paper?}

\textbf{Comparative Study: }To evaluate the effectiveness of MSPRP, we compare baseline models with their MSPRP-enhanced models. The results show clear and consistent improvements across all assessed dimensions.

In terms of basic metrics, MSPRP increases BLEU-n, ROUGE-L, METEOR, and BERTScore for both small and large models (e.g., Qwen2.5-7B and Qwen2.5-72B). These gains suggest that MSPRP helps models generate answers that more closely match real patient responses, reflecting improvements in content alignment and semantic fidelity.

The improvements are more striking in semantic evaluations. MSPRP brings substantial enhancements in persona consistency, factual consistency, naturalness, and contextual relevance. Notably, persona consistency improves the most, confirming that the multi-stage regulation mechanism effectively stabilizes the projection of the target persona across dialogue turns. Moreover, improvements in naturalness indicate that the communication-style and expression-regulation stages lead to linguistic behaviors that better resemble real patients.

\textbf{Ablation Study: }To further validate the contribution of the components of the MSPRP framework, we conducted ablation experiments aimed at answering two core questions: (1) the effectiveness of each stage, and (2) the impact of the execution sequence. Tables \ref{tab:ablation_llm_metrics1} and \ref{tab:ablation_llm_human_eval1} show the results of automated and manual evaluations, respectively. The single-stage application outperformed the baseline on all metrics, confirming the independent value of each module. Among them, Stage 1 (basic information generation) significantly improved performance at the basic evaluation, and Stage 2 (communication style injection) achieved the most obvious optimization in the persona evaluation. In the multi-stage combination experiment, the results of Stage 2+3+1 confirm that the factual accuracy guaranteed by Stage 1 is the basic prerequisite for the effective operation of the entire framework. The performance of Stage 1+3+2 confirms the patient’s expression logic, showing that a gradual progression from internal style shaping to external language output aligns better with real-world communication needs. Stage 1+2+3 performs the best, outperforming other combinations in both basic evaluation metrics and persona evaluation metrics.

In summary, the effectiveness of the MSPRP framework stems from three core factors: the inherent value of each independent stage, the complementary gains generated from the combination of multiple stages, and the progressive adjustment process of ``factual basis→ style shaping→ expression refinement'', which provides key support for optimizing the quality of language generation and the fidelity of persona.

\begin{table}
\centering
\caption{Ablation results of each stage and stage ordering in MSPRP on basic evaluation. Baseline: Qwen2.5-7B. Stage 1: Basic Information Generation, Stage 2: Communication Style Injection, and Stage 3: Expression Consistency Regulation. }
\vspace{3pt}
\resizebox{\textwidth}{!}{
\begin{tabular}{lccccccc}
\hline
Model & BLEU-1 & BLEU-2 & BLEU-3 & BLEU-4 & ROUGE-L & METEOR & BERTScore \\
\hline
Baseline    & 0.1828 & 0.0856 & 0.0504 & 0.0329 & 0.2051  & 0.2004 & 0.6437    \\
    Stage 1       & 0.1999 & 0.0937 & 0.0552 & 0.0360 & 0.2177  & 0.2004 & 0.6483    \\
    Stage 2       & 0.1904 & 0.0884 & 0.0518 & 0.0339 & 0.2110  & 0.2002 & 0.6457    \\
    Stage 3       & 0.1929 & 0.0900 & 0.0530 & 0.0348 & 0.2126  & 0.1997 & 0.6457    \\
    Stage2+3+1    & 0.1936 & 0.0900 & 0.0526 & 0.0342 & 0.2115  & 0.2019 & 0.6459    \\
    Stage1+3+2    & 0.1934 & 0.0897 & 0.0522 & 0.0338 & 0.2129  & 0.2050 & 0.6467    \\
    Stage1+2+3    & 0.2031 & 0.0956 & 0.0569 & 0.0375 & 0.2196  & 0.2041 & 0.6491    \\
\hline
\end{tabular}
}
\label{tab:ablation_llm_metrics1}
\end{table}

\begin{table}
\centering
\caption{Ablation results of each stage and stage ordering in MSPRP on persona evaluation.}
\vspace{3pt}
\resizebox{\textwidth}{!}{%
\begin{tabular}{lcccc}
\hline
Model & Persona Consistency & Factual Consistency & Naturalness & Contextual Relevance \\
\hline
 Baseline    & 3.748 & 3.795 & 3.824 & 3.780 \\
    Stage 1       & 3.777 & 3.818 & 3.846 & 3.835 \\
    Stage 2       & 3.821 & 3.835 & 3.882 & 3.849 \\
    Stage 3       & 3.786 & 3.828 & 3.859 & 3.842 \\
    Stage2+3+1    & 3.816 & 3.846 & 3.876 & 3.860 \\
    Stage1+3+2    & 3.836 & 3.843 & 3.886 & 3.862 \\
    Stage1+2+3    & 3.905 & 3.918 & 3.942 & 3.937 \\
\hline
\end{tabular}%
}
\label{tab:ablation_llm_human_eval1}
\end{table}

These results validate that MSPRP is an effective, training-free solution for enhancing patient-simulation quality. By decomposing generation into structured stages and applying persona regulation, the framework compensates for LLMs’ inherent tendency toward generic and uniform responses.

\begin{figure*}[t]
  \centering
  \includegraphics[width=\textwidth]{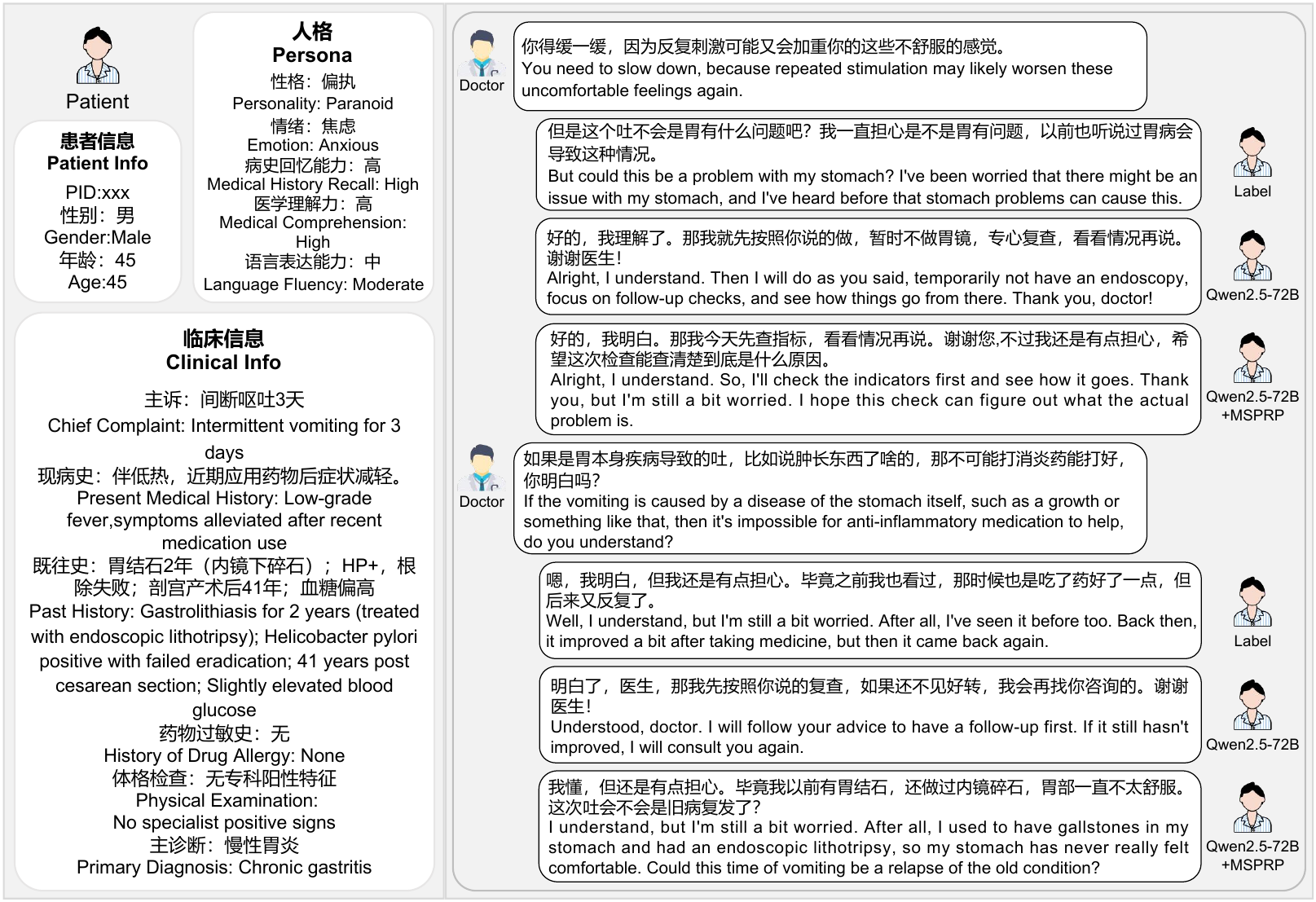} 
  \caption{Case Study. Comparison of Original Responses, Basic Responses, and Responses Applying the MSPRP Framework to Physician Questions.} 
  \label{ffirst} 
\end{figure*}

\subsection{RQ3: Are LLMs’ patient-simulation behaviors natural and clinically plausible in realistic doctor–patient communication scenarios?}

To address RQ3, we analyze the naturalness and clinical plausibility of LLM patient simulations through quantitative metrics and qualitative cases, with results varying significantly between models with and without the MSPRP framework.

Quantitatively, MSPRP-enhanced models show notable improvements in key dimensions. As seen in Table \ref{tab:llm_human_eval1}, Qwen2.5-72B+MSPRP scores 3.970 in Naturalness and 3.969 in Contextual Relevance, up from the baseline’s 3.914 and 3.910, while its Factual Consistency reaches 3.956, ensuring responses are clinically accurate and contextually coherent. 

Qualitative observations further validate the effectiveness of the MSPRP framework. As shown in Figure \ref{ffirst}, when faced with doctors' diagnostic questions, the base model can provide factually correct answers but tends to be overly formal, lacks emotional expression, and fails to demonstrate the retrieval and response to relevant medical information related to the patient. In contrast, the output of the model enhanced by MSPRP is more aligned with the linguistic characteristics of patients in real clinical settings. For example, when a doctor suggests postponing a gastroscopy to alleviate the patient's physical discomfort, the simulated anxious-type patient under the MSPRP framework will proactively express concerns about ``postponing the examination may not clarify the cause''. Even after the doctor provides a secondary explanation, the patient will still reiterate their concerns based on their paranoid personality traits. The response also combines the patient's medical history to substantiate the validity of his concerns, which aligns with the ``high medical history recall ability'' dimension and confirms the value of the MSPRP framework in Stage 1.

Overall, both quantitative and qualitative results indicate that MSPRP substantially enhances the naturalness and clinical plausibility of simulated patient interactions, bringing model behavior closer to real-world doctor–patient communication patterns.

\section{Conclusion}
In this paper, we investigate the simulation of realistic patient behavior as a crucial component for advancing clinical LLMs and enhancing medical diagnostic education. We present Ch-PatientSim, the first Chinese patient simulation dataset constructed from authentic clinical interaction scenarios, enabling a comprehensive evaluation of LLMs’ ability to emulate patient traits. By modeling patients based on a five-dimension persona structure and augmenting the dataset via few-shot generation with LLMs followed by manual verification, we ensure diversity and realism. Our evaluation of state-of-the-art LLMs reveals that many models generate responses that are overly formal and lack individualized personality. To overcome this limitation, we introduce a training-free Multi-Stage Patient Role-Playing (MSPRP) framework, which decomposes patient-LM interactions into three stages to maintain both personalization and contextual realism. Experimental results show that MSPRP substantially enhances model performance across multiple dimensions of patient simulation, including persona consistency, factual accuracy, naturalness, and contextual relevance. These findings underscore the importance of structured role-playing frameworks in improving LLM-based patient simulations.


\section{Acknowledgements}

This work is supported by the National Natural Science Foundation of China [62576149] and the Fundamental Research Funds for the Central University, JLU.

 \bibliographystyle{elsarticle-harv} 
 \bibliography{sample}

\end{document}